\documentclass{article} 
\pdfoutput=1 

\usepackage{iclr2020_conference,times}

\usepackage{amsmath,amsfonts,bm}

\def\eqref#1{equation~\ref{#1}}

\def\1{\bm{1}}

\DeclareMathAlphabet{\mathsfit}{\encodingdefault}{\sfdefault}{m}{sl}
\SetMathAlphabet{\mathsfit}{bold}{\encodingdefault}{\sfdefault}{bx}{n}

\usepackage{hyperref}
\usepackage{url}
\usepackage{amsthm}
\usepackage{amsmath}
\usepackage{amsfonts}
\usepackage{ragged2e}
\usepackage{bbm}

\usepackage{float}
\usepackage{subcaption}
\usepackage[labelformat=parens,labelsep=quad,skip=3pt]{caption}
\usepackage{graphicx}

\newtheorem{defi}{Definition}
\newtheorem{thm}{Theorem}
\newtheorem{lemma}{Lemma}

\let\vec\mathbf

\newcommand{\x}{{\vec x}}
\newcommand{\y}{{\vec y}}
\newcommand{\z}{{\vec z}}

\title{ReLU Code Space: A Basis for Rating Network Quality Besides Accuracy}

\author{Natalia Shepeleva, Werner Zellinger, Michal Lewandowski and Bernhard Moser \\
Data Analysis Systems\\
Software Competence Center Hagenberg, Austria\\
\texttt{\{natalia.shepeleva, werner.zellinger, michal.lewandowski,}\\
\texttt{ bernhard.moser\}@scch.at}}

\iclrfinalcopy 
\begin{document}

\maketitle

\begin{abstract}
We propose a new metric space of ReLU activation codes equipped with a truncated Hamming distance which establishes an isometry between its elements and polyhedral bodies in the input space which have recently been shown to be strongly related to safety, robustness, and confidence.
This isometry allows the efficient computation of adjacency relations between the polyhedral bodies.
Experiments on MNIST and CIFAR-10 indicate that information besides accuracy might be stored in the code space. \footnote{Code available at: \url{https://github.com/nataliaShepeleva/ReLU_Code_Space_NAS-ICLR2020}}
\end{abstract}

\section{Motivation}

In this work, we propose a new metric space as a basis for quality indices used to rate quality aspects of ReLU networks.
Such quality indices are often used to choose between networks with similar accuracy.
For example, quality indices like the number of parameters and the training time are used to choose between models in neural architecture search, see e.g.~\cite{ying2019bench}.
Other recent examples are the fine-grained loss and the fine-grained accuracy which are also applied in neural architecture search~\cite{dong2020bench}.

Our work is inspired by~\cite{Montufar2014b} who analyses the complexity of ReLU networks  in terms of the number of regions on which
the ReLU is linear. These regions turn out to be finite intersections of halfspaces. 

Recursion formulas for explicitly computing the resulting polyhedral bodies were introduced recently by~\cite{MoserPatent2018} and, slightly later, by~\cite{Croce2018}.
Both works relate the polyhedral bodies to safety and robustness by utilizing the recursion formulas for strategies against adversarial attacks. 
In particular, the work of~\cite{Croce2018} indicates that larger, and consequently a lower number of polyhedral bodies induce a higher robustness of ReLU networks.

\cite{Hein2018}~point out that unjustified high confidence of some networks can be explained by the unboundedness of polyhedral bodies in outer regions.
In unbounded polyhedral bodies there are rays into infinity along which Softmax values converge to $1$ pretending high confidence independently from whether this confidence is justified or not.
See also~\cite{Croce2019a} for a robustness analysis of ReLU networks under adversarial attacks based on this approach.
In \cite{Croce2019} the polyhedral structure is exploited to derive provable robustness against certain types of adversarial attacks.
See also~\cite{jordan2019provable} for a similar approach. 

In contrast, our work provides a different algebraically motivated approach to the polyhedral bodies.
We start our analysis by establishing the equivalence relation between points in the input space where points are equivalent if and only if they show the same binarized activation behavior. This approach leads to a refined analysis 
with main contributions as follows:
\begin{itemize}
 \item We propose a new metric space (ReLU code space) in which ReLU network-induced polyhedral bodies and their adjacency relationships can be efficiently represented and computed.
\item We give an abstract geometric and algebraic characterisation of the new metric space.
\item We perform experiments on MNIST and CIFAR-10 indicating that information besides accuracy might be stored in the code space, which show future potential of our method for NAS systems.
\end{itemize}

This work is structured as follows: Section~\ref{sec:code_space} describes our approach, Section~\ref{sec:experiments} gives our experiments including discussions, and Appendix~\ref{appendix_A} summarizes the required background and gives the proof of our main Theorem~\ref{thm:duality}.

\section{New metric space for quality indices}
\label{sec:code_space}
In the following let $f:\mathbb{R}^m\to \mathbb{R}^n$ be a ReLU network, namely
     $f( 
    \x )= \xi \circ \mathrm{relu}\circ g_{l} \circ \ldots \circ \mathrm{relu}\circ g_1(\x) $ with $g_k(\x)$ being parametric affine functions for $k \in \{1,\ldots,l\}$
as in Definition \ref{def:neural_network}. 
Further denote by $N=\sum_{k=1}^l n_k$ the total number of neurons and by ${\vec a}_{k}(\x)=(a_{1}^{(k)}(\x),\ldots,a_{n_k}^{(k)}(\x))=\mathrm{relu}\circ g_{k}\circ \ldots \circ \mathrm{relu}\circ g_0(\x)$ the activation vector of some input $\x\in\mathbb{R}^m$ at layer $k$.
Given some input sample $\{\x_1,\ldots,\x_s\}$ with elements being realizations of iid random variables, the learning a ReLU network $f$ means to find the unknown parameters.
\begin{defi}
We define the  {\bf ReLU-code space} $(\mathcal{X}_f,d_\mathrm{H, \theta})$ of a ReLU network $f:\mathbb{R}^{m}\to\mathbb{R}^{n}$ as 
metric space consisting of the set of induced codes, i.e.,
\begin{align}
    \mathcal{X}_f=\left\{\mathrm{code}_f(\x)\in\{0,1\}^N\,\middle|\, \x\in\mathbb{R}^{n}\right\}
\end{align}
and as metric the {\bf truncated Hamming distance} $d_\mathrm{H,\theta}(a,b)=\min\{d_\mathrm{H}(a,b), \theta\}$
with threshold $\theta \in \mathbb{N}\cup \{\infty\}$ and $d_\mathrm{H}(a,b)=|\{j\in\{1,\ldots,N\}\mid a_j\neq b_j\}|$
between the {\bf codes} $a=\mathrm{code}_f(\x)$, $b=\mathrm{code}_f(\y)$ induced by the input vectors $\x,\y\in \mathbb{R}^{m}$, where
\begin{align}
\begin{split}
\label{eq:code}
    \mathrm{code}_f:
    \x \mapsto (\beta_{1}^{(1)},\ldots,\beta_{n_1}^{(1)},\ldots,\beta_{1}^{(n_l)},\ldots,\beta_{n_l}^{(n_l)})\in \{0,1\}^N
\end{split}
\end{align}
with $\beta_{i}^{(k)}=1$ if $a_i^{(k)}(x)>0$ and 
$\beta_{i}^{(k)}=0$ else.
\end{defi}
By the definition above, the codes $\mathrm{code}_f(\x)$ and $\mathrm{code}_f(\y)$ of two different points $\x,\y\in\mathbb{R}^m$ are the same if the ReLU network $f$ assigns the same partial linearity to $\x$ and $\y$.

In the following, we characterize the subset of all points in the input space, which yield the same code, see Appendix~\ref{appendix_A} for its proof.
\begin{thm}[{\bf Duality Representation Theorem of ReLU  Codes}]
\label{thm:duality}
    With the equivalence relation
    \begin{align}
        \label{eq:equiv}
        \x\sim_f \y :\!\iff d_\mathrm{H}(\mathrm{code}_f(\x),\mathrm{code}_f(\y))=0
    \end{align} and equivalence classes
     \begin{align}
        \label{eq:equivClass}
        [\x]_f := \{{\bf z} \in \mathbb{R}^m|\, {\bf z}\sim_f \x \}
    \end{align}
    the following holds:
    \begin{enumerate}
        \item There is a one-to-one correspondence between the code space $\mathcal{X}_f$ and the set of equivalence classes $\{[\x]_f|\, x \in \mathbb{R}^m\}$. 
        \item The topological closure $\overline{[\vec x]}_f$ of the equivalence class $[\vec x]_f$ is a polyhedron, i.e., the intersection of a finite number of closed half-spaces.
        \item The equivalence class $[\vec x]_f$ is the (disjoint) union of relative interiors $relint(F)$ of subfaces $F$ of the polyhedron $P =\overline{[\vec x]}_f$ forming a lattice structure.
    \end{enumerate}
\end{thm}
To get further intuitions about the topological properties induced by the Hamming distance on ReLU codes, we now give a result which directly follows from Corollary 4.3 in~\cite{jordan2019provable}.
\begin{lemma}[{\bf Adjacency Lemma}]
    \label{lemma:croce}
    {\justifying
    Let $f$ be a ReLU network with parameters in general position. Then $d_\mathrm{H}(\mathrm{code}_f(\x),\mathrm{code}_f(\y))=1$ iff the polyhedral bodies $\overline{[\x]}_f$ and $\overline{[\y]}_f$ as defined by Equation~(\ref{eq:equiv}) are \textbf{adjacent}, i.e.,~$\mathrm{dim}(\overline{[\x]}_f\cap \overline{[\y]}_f)=\mathrm{dim}(\overline{[\x]}_f)-1=\mathrm{dim}(\overline{[\y]}_f)-1$.}
\end{lemma}

Lemma~\ref{lemma:croce} tells us that the adjacency relation between polyhedral cells $[x]_f$ and $[y]_f$ is reflected by the truncated Hamming distance $d_{H,2}$ of the corresponding codes $code_f(x)$ and $code_f(y)$. 
To this end, we obtain an isometry between the input space and the code space by means of the following definition.

\begin{defi} Given a ReLU network $f:\mathbb{R}^m \rightarrow \mathbb{R}^n$ and its induced tessellation 
$\tau = \{[\x]_f\mid \x \in \mathbb{R}^m \}$, we define the {\bf adjacency metric space} induced by $\tau$ by $(\tau, d_{\cal{A}})$, where 
\begin{equation}
  d_{\cal{A}}(\x,\y):= d_{H,2}(code_f(\x),code_f(\y)) \in \{0,1,2\}.
\end{equation}
\end{defi}

An adjacency distance $0$ means that $x\sim_f y$, a distance $1$ means that the corresponding polyhedral cells $\overline{[x]}_f$ and $\overline{[y]}_f$ are distinct but adjacent (sharing a common subface of dimension $m-1$), and $2$ means that the cells are distinct and not adjacent. Note that a Hamming distance larger than one between two codes relates to a more complex neighbourhood relation between the corresponding polyhedral bodies.

Summing up, we obtain Theorem~\ref{thm:isometry}.
\begin{thm}[{\bf ReLU Code Space Isometry Theorem}]
\label{thm:isometry}
The mapping $code_f$ of Equation~(\ref{eq:code}) establishes an isometry between 
$(\tau, d_{\cal{A}})$ and $(\mathcal{X}_f, d_{H,2})$.
\end{thm}

In Theorem~\ref{thm:duality} and Theorem~\ref{thm:isometry}  we provide the interpretation of $\mathrm{code}_f(\x)$ as representation of the equivalence class $[\x]_f$ in the code space but we also clarify the geometric structure of the equivalence classes $[\vec x]_f \subset \mathbb{R}^n$ beyond its interpretation as a sub-region of linear activation functions as used by~\cite{Montufar2014b,Croce2019}, respectively. 

To this end, the subtle topological analysis in Theorem~\ref{thm:duality} whether border points belong to
an equivalence class $[x]_f$ turns out to be the key for revealing the fundamental property of isometry of Theorem~\ref{thm:isometry}. 
Due to Theorem~\ref{thm:isometry} we may use synonymously 
{\it clustering in the  ReLU code space} and {\it clustering at the cell-level} of the induced tessellation, in short {\it cell-level clustering}.


\section{Experiments}
\label{sec:experiments}
As indicated by Theorem~\ref{thm:isometry}, the binarization of ReLU activation values allows the efficient analysis of the underlying 
tessellation. First experiments underpin our conjecture that characteristics of the tessellation such as the number of non-empty cells (containing a training point) are informative indicators for analyzing the behavior of 
neural networks besides accuracy. This means that characteristics of the tessellation and their correspondences in the ReLU code space can be helpful for deriving novel quality measures.

\begin{figure}[htb]
    \centering 
\begin{subfigure}{0.5\textwidth}
  \includegraphics[width=\linewidth]{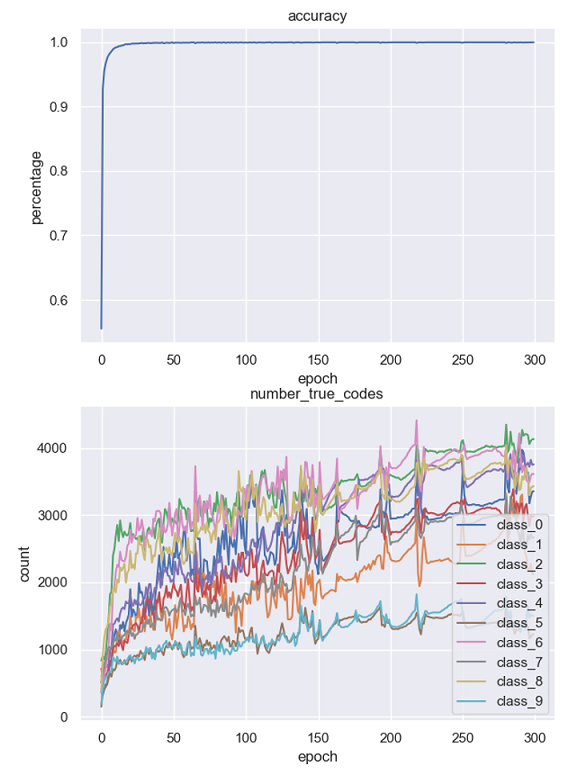}
 \caption{learning rate 0.001}
  \label{fig:1}
\end{subfigure}\hfil 
\begin{subfigure}{0.49\textwidth}
  \includegraphics[width=\linewidth]{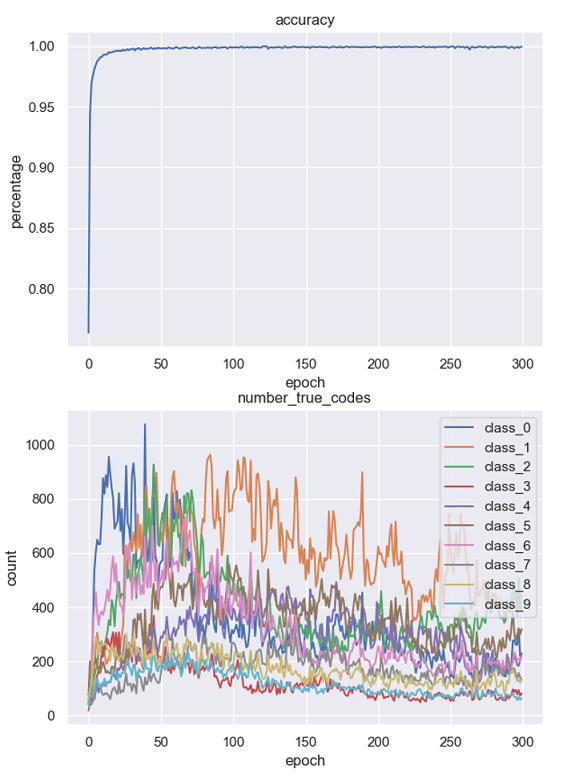}
 \caption{learning rate 0.01}
  \label{fig:2}
\end{subfigure}\hfil 
\caption{VGG16 network trained on MNIST with different learning rates, a) 0.001 and b) 0.01. Both networks are indistinguishable with respect to  accuracy. However, the number of cells of the induced tessellation with correctly classified points shows a different behavior.}
\label{fig:acc_diff}
\end{figure}

In our {\bf first experiment}, see Figure~\ref{fig:acc_diff}, we trained a VGG16 network on MNIST with two different learning rates. Although the accuracy at the end of both training procedures is not distinguishable, the number of codes, i.e., the number of non-empty cells of the induced tessellation, evolves noticeably differently over the epochs of training.  
This underpins the interpretation that Theorem~\ref{thm:isometry} provides the right abstraction level to capture and to reveal
topological structures such as connectedness (with respect to the adjacency relation) at the cell level of the induced tessellation.

\begin{figure}[htb]
    \centering 
\begin{subfigure}{0.5\textwidth}
  \includegraphics[width=\linewidth]{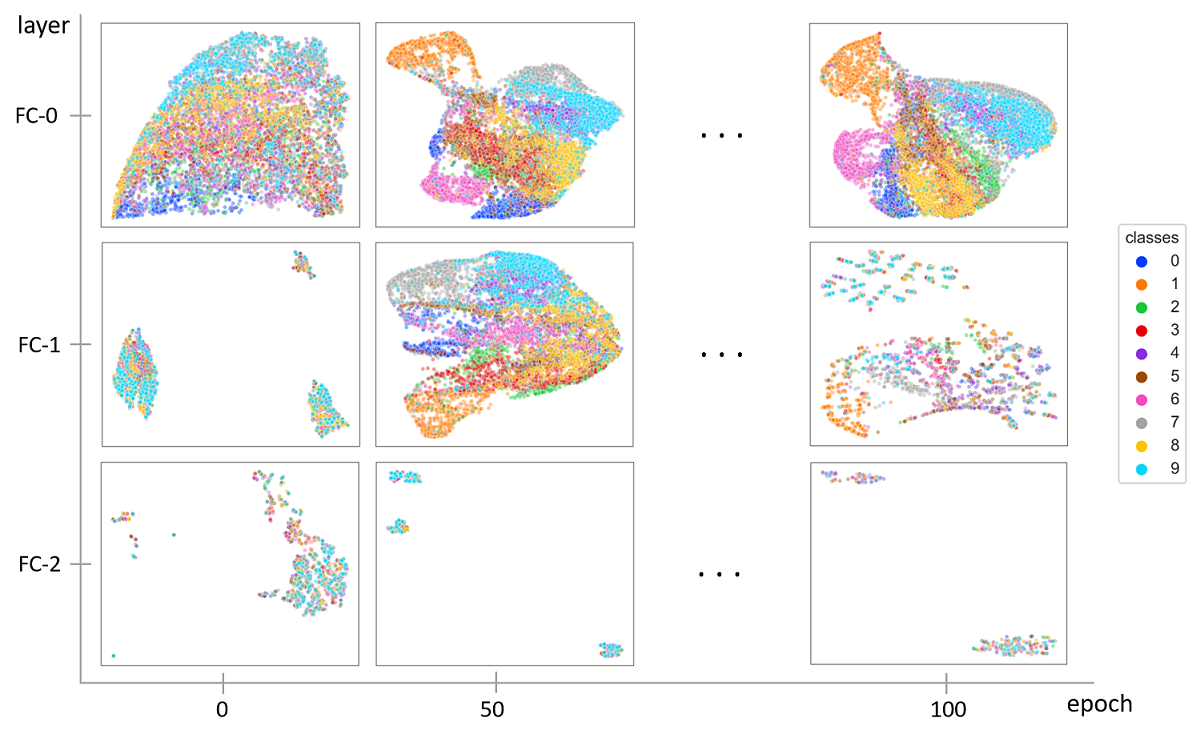}
  \caption{Autoencoder}
  \label{fig:ae}
\end{subfigure}\hfil 
\begin{subfigure}{0.5\textwidth}
  \includegraphics[width=\linewidth]{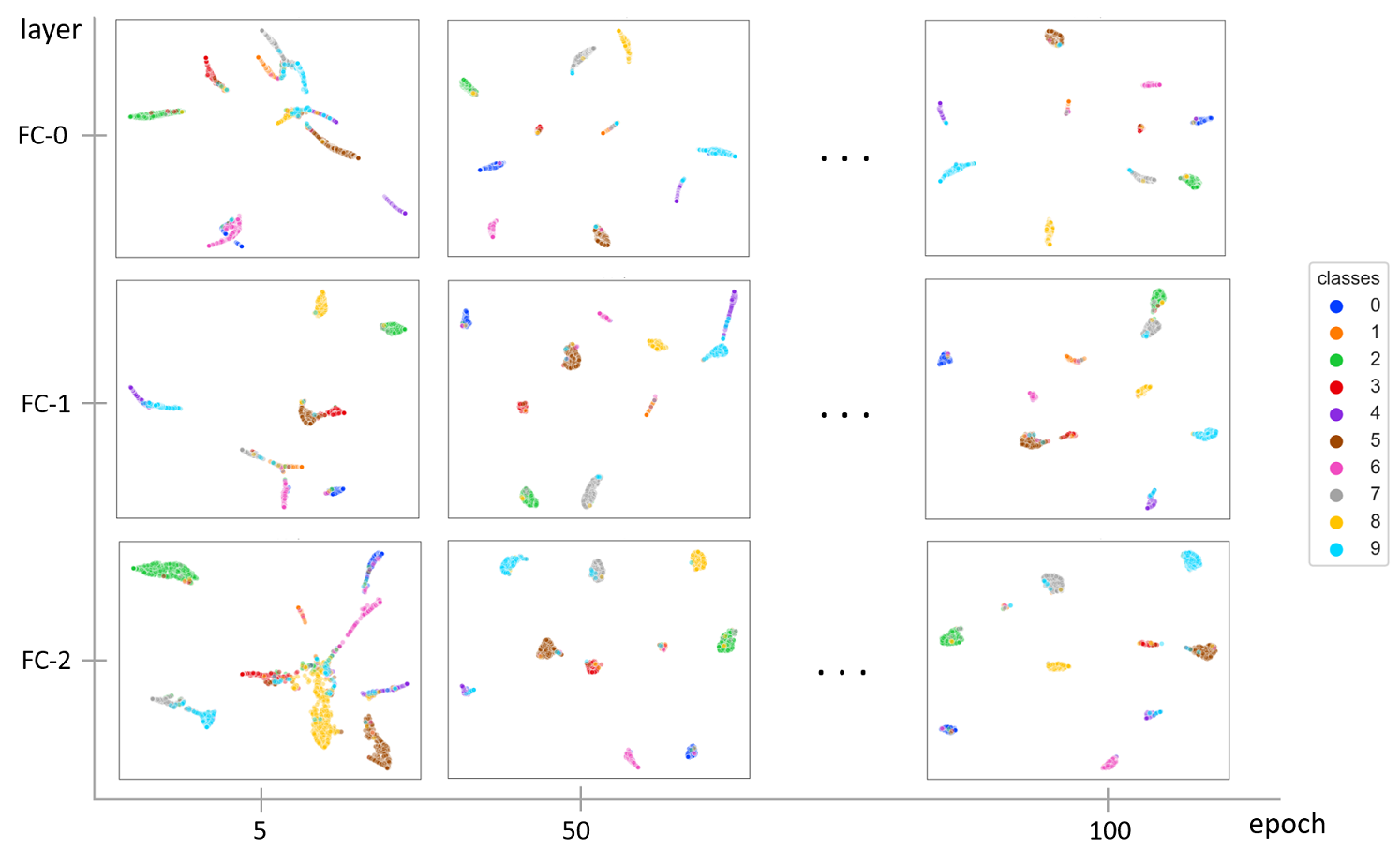}
  \caption{VGG16}
  \label{fig:vgg16}
\end{subfigure}\hfil 
\caption{Autoencoder and VGG16 network trained on MNIST with learning rate 0.01. Rows show changes of the clusters of ReLU codes over the training time, columns show corresponding changes within the network layers.}
\label{fig:cluster}
\end{figure}

This interpretation is further underpinned by our {\bf second experiment} in a clustering setting, see Figure~\ref{fig:cluster}.
We consider two settings: autoencoder, see Figure~\ref{fig:ae},  and classification, see
Figure~\ref{fig:vgg16}.
Regarding Figure~\ref{fig:ae}, we trained a three-layered autoencoder on MNIST 
and embedded its ReLU-codes into a two dimensional space using the  dimension reduction technique UMAP~\cite{mcinnes2018umap}. The visible clusters of ReLU codes for different classes indicate, as expected, discriminative information while in higher layers clusters of ReLU codes of different classes cannot be well separated.
As training successfully progresses, this effect gets stronger.
Therefore, information about the expected behaviour of the autoencoder is presented in the ReLU code space, as features of higher layers are expected to learn more invariant features than lower ones.
Analogously, the cell-level clusters behave as expected in a classification setting as shown in Figure\ref{fig:vgg16}.
We trained a classifier (VGG16) on the same datasets.
It can be seen that higher layers store more discriminative information than lower ones in this case. This behaviour is as expected, see e.g.~\cite{alain2016understanding} and indicates that binarized ReLU codes contain the essence of information for discrimination.

Further experiments and plots based on CIFAR-10 can be found in the Appendix~\ref{appendix_B}.

\section{Conclusion}
\label{sec:conclusion}
Our paper provides the {\bf starting point} for more complex geometric analysis in the input space such as exploring the geometry of manifolds of points belonging to the same class of objects by means of discrete binary operations in the code space. Above all the {\bf ReLU Code Space Isometry Theorem}~\ref{thm:isometry} allows the efficient computation of the adjacency relation which boils down to checking the Hamming distance of binarized activation states. To this end, we showed that the binarization of ReLU activation values is useful for several reasons: a) establishing an isometry that allows the efficient representation of the ReLU network-induced tessellation of the input space and the efficient computation of adjacency relations between its cells, b) by this, providing a tool to analyse ReLU networks beyond accuracy, as indicated by first experiments on MNIST.

\section*{Acknowledgements}
The research reported in this paper has been partly funded by BMVIT, BMDW, and the Province of Upper Austria in the frame of the COMET Programme managed by FFG and the COMET module S3AI No. 872172 as well as from the European Union’s Horizon 2020 Research and Innovation programme under grant agreement No. 780788.


\bibliography{iclr2020_conference}
\bibliographystyle{iclr2020_conference}

\appendix
\section{Appendix}
\label{appendix_A}
In this subsection we define a ReLU-based neural network we work with in this work.
\begin{defi}
[ReLU Network, see e.g.~\cite{Goodfellow-et-al-2016,berner2018analysis}]%
\label{def:neural_network}%
A ReLU network $f$ is a function
\begin{align}
    \begin{split}
    f:\mathbb{R}^m \ni 
    \x \mapsto \xi \circ \mathrm{relu}\circ g_{l} \circ \ldots \circ \mathrm{relu}\circ g_1(\x) \in \mathbb{R}^{n}
    \end{split}
\end{align}
with $l\in\mathbb{N}$ (hidden) layers, element-wise application of  $\mathrm{relu}: \mathbb{R}\ni z\mapsto \max\{0,z\}\in \mathbb{R}$,  and output function $\xi$. For $k\in\{0,\ldots,l\}$ and $n_0=m$, the linear functions $g_k$ are defined by
\begin{align}
    \begin{split}
        g_k:\mathbb{R}^{n_k} \ni 
        \x \mapsto {\vec W}_k \x + {\vec b}_k \in \to\mathbb{R}^{n_{k+1}} \, \mathrm{with}\, {\vec W}_{k+1}\in\mathbb{R}^{n_{k+1}\times n_{k}}\, \mathrm{and}\, {\vec b}_{k}\in\mathbb{R}^{n_{k}}.
    \end{split}
\end{align}
The number of neurons is defined by $N=\sum_{k=1}^l n_k$ and the activation vector of some input $\x\in\mathbb{R}^m$ at layer $k$ is given by ${\vec a}_{k}(\x)=(a_{1}^{(k)}(\x),\ldots,a_{n_k}^{(k)}(\x))=\mathrm{relu}\circ g_{k}\circ \ldots \circ \mathrm{relu}\circ g_0(\x)$.
\end{defi}
\subsection{Preliminaries on Polyhedra and Lattices}
\label{ss:AbstractPolytopes}

The terms \textit{polytope} and \textit{polyhedron} are not consistently used in the literature. 
We stick to the references~\cite{brondsted2012introduction,mcmullen1973representations} and shortly recall their definition. 
A {\it polytope} is the convex hull of a finite number of points. 
A polyhedron or polyhedral set $P$ is the intersection of a finite number of closed halfspaces or $P$ equals the whole space, i.e.~it is given as set $\{\vec x\in \mathbb{R}^m\mid\, A\, \vec x \leq \vec b\}$.
A bounded polyhedron is a polytope.

Further, we make a distinction between a polyhedron and a {\it polyhedral body}. 
In contrast to a polyhedron a polyhedral body results from the intersection of finitely many either closed or open halfspaces.
  
In abstract algebra a {\it lattice} is a pair $\mathcal{L}=(L,\prec)$ consisting of a set $L$ and a partial order relation 
$\prec$ for  which every two elements have a unique {\it least upper bound} and a unique {\it greatest lower bound}. 
An example is given by the set of subfaces $F$ of a polytope together with the emptyset $\emptyset$ and the set inclusion as partial order relation. This lattice is closely related to the notion of abstract polytopes, see~\cite{mcmullen1973representations}.

In this context we also use notation from point set topology. For a set $A \subseteq \mathbb{R}^k$, 
the interior $A^{\circ}$ of $A$ is the set of all points $x \in A$ for which an Euclidean ball 
$B_{\epsilon}(x)$ with center at $x$ and radius $\epsilon>0$ is contained in $A$. The closure $\overline{A}$ of $A$ is the set of all points $x \in \mathbb{R}^k$ for which all (non-empty) balls $B_{\epsilon}(x)$ have a non-empty intersection with $A$. 
For example, the closure of a polyhedral body is a polyhedron.
The {\it relative interior} $relint(A)$ of a set $A$ as subset of a hyperplane (or subspace of $\mathbb{R}^m$) is its interior w.r.t.~the relative topology restricted to the subspace.
For any nonempty convex set $F \subseteq \mathbb{R}^m$ the relative interior can be characterized as
\begin{equation}
\label{eq:relint}
relint(F) = \{\vec x \in F\mid\, \forall \vec y \in F\, \exists \lambda >0: \lambda \vec x +(1- \lambda) \vec y \in F\}.
\end{equation}
Note that the relative interior of a single point is the point itself, i.e.~$relint(\{\vec p\})=\{\vec p\}$, and that the relative interior of a straight line with endpoints $\vec a$ and $\vec b$ is the set of all points of that line except $\vec a$, $\vec b$.

\subsection{First Part of Proof of Theorem~\ref{thm:duality}}
\label{ss:first_part_of_proof}

As $\{[\vec x]_f\mid  \vec x \in \mathbb{R}^m\}$ is a partition of $\mathbb{R}^m$ (defining property of the equivalence relation), further, $[\vec x]_f$ only depends on $code(\vec x) \in \{0,1\}^N$, and
$code_f(\x_1)\neq  code_f(\x_2)$ implies $[\x_1]_f \cap [\x_2]_f = \emptyset$, the mapping 
\begin{equation}
\label{eq:one-to-one}
\gamma: \{[\vec x]_f \subseteq \mathbb{R}^m\mid  \vec x \in \mathbb{R}^m\} \rightarrow \mathcal{X}_f=\{code_f(\vec x) \in \{0,1\}^N\in \mid \vec x\in \mathbb{R}^m\}
\end{equation}
is a one-to-one mapping. 
That is, there is a one-to-one correspondence between binary sequences (neural codes)
and equivalence classes $[\x]_f$.\hfill\qed

\subsection{Second Part of Proof of Theorem~\ref{thm:duality}}
\label{ss:second_part_of_proof}
We will recall the approach of~\cite{MoserPatent2018}.
Consider the 'binary activation states' 
\begin{equation}
\label{eq:binarystate}
\beta_{i_k}^{(k)}=
    \begin{cases}
    1:~a_{i_k}^{(k)}(\x)>0\\
    0:~\text{else}
\end{cases}
\end{equation}
and the corresponding 'polar activation states' 
\begin{equation}
\label{eq:polarstate}
\pi^{(k)}_{i_k} := 2\, \beta^{(k)}_{i_k} -1 \in \{-1,1\}.
\end{equation}
By this we define the diagonal (activation profile) matrices 
\begin{equation}
\label{eq:Q}
Q^{(k)}_{\pi}  := \mbox{Diag}[\pi^{(k)}_1, \ldots, \pi^{(k)}_{n_k}],\,
Q^{(k)}_{\beta} :=  \mbox{Diag}[\beta^{(k)}_1, \ldots, \beta^{(k)}_{n_k}],
\end{equation}
where the index $k$ refers to the layer.

Now, consider a network with only one layer, i.e., $l=1$, and multiply each row $v_i$ of the vector
$(v_1, \ldots, v_{n_1})^T:=(W_1\, \widetilde \x + \vec b_1)$ by $+1$ if $relu(v_i)>0$
and by $-1$ if $relu(v_i)=0$, i.e., $v_i\leq 0$. 
As a result we get a vector with non-negative entries, namely 
$Q^{(1)}_{\pi}(W_1\, \widetilde \x + \vec b_1) \geq 0$. 
As the diagonal of $Q^{(1)}_{\pi}$ refers to the polar representation of the activation profile induced by
$\widetilde \x$, it follows that a point $\widetilde \x$ with this activation profile is element of the polyhedron 
\begin{equation}
\label{eq:P}
P_{\widetilde \x} := \{\vec x\in \mathbb{R}^m\mid\, \, W_{\widetilde \x}^{(1)}\vec x + \vec b_{\widetilde \x}^{(1)} \geq 0\},
\end{equation}
where 
\begin{equation}
\label{eq:recursion1}
W_{\widetilde \x}^{(1)} :=  Q^{(1)}_{\pi}\,W_1,\quad \vec b_{\widetilde \x}^{(1)}:= Q^{(1)}_{\pi}\,\vec b_1
\end{equation}
Consequently, we have 
\begin{equation}
    \label{eq:classPinclusion}
    [\widetilde \x]_f \subseteq P_{\widetilde \x}.
\end{equation}

This idea of switching the signs of the pre-activation according to whether the activation is positive or not can recursively be applied to subsequent layers. Note that for an intermediate layer $k< l$ the output of a neuronal unit to the next layer is $0$ if there is no activation. 
Therefore, instead of using the polar activation profile matrix $Q_{\pi}^{(k)}$ we have to apply its binary variant 
$Q_{\beta}^{(k)}$.
So, for a network of $2$ layers $(W_1, b_1; W_2, b_2)$ we obtain: 
\[
Q_{\pi}^{(2)}\left(W_2 \left[Q_{\beta}^{(1)} (W_2 \widetilde \x + \vec b_1) \right] + \vec b_2\right) \geq 0.
\]
Expanding the left hand side of this inequality leads to an equivalent representation of a polyhedron in the style 
of~(\ref{eq:P}). 

In general, given a ReLU network $(W_j, \vec b_j)_{j=1, \ldots,L}$ with $l$ layers and the input data $\widetilde \x$, it turns out that 
$[\widetilde \x] \subseteq P_{\widetilde \x}$, where  $P_{\widetilde \x}$ is a polyhedron given by 
\begin{eqnarray}
\label{eq:eqclass}
W_{\widetilde \x}^{(k)} \vec x + \vec b_{\widetilde \x}^{(k)} \geq 0, \,\, k = 1, \ldots, L,
\end{eqnarray}
where
\begin{eqnarray}
\label{eq:recursion}
W_{\widetilde \x}^{(k)} &:=& Q_{\pi}^{(k)} W_k \prod_{j=1}^{k-1} Q_{\beta}^{(k-j)} W_{k-j}, \nonumber \\
\vec b_{\widetilde \x}^{(k)} &:=& 
Q_{\pi}^{(k)} W_k 
\sum_{i=1}^{k-1} 
\left( \prod_{j=1}^{k-i-1} 
Q_{\beta}^{(k-j)}  W_{k-j}
\right) Q_{\beta}^{(i)}  \vec b_i
+ 
Q_{\pi}^{(k)} \vec b_k.
\end{eqnarray}

Note that the derivation for (\ref{eq:eqclass}) analogously can by applied to the interior 
$[\widetilde \x]_f^{\circ}$ of $[\widetilde \x]_f$ showing the 
equivalence
\begin{eqnarray}
\label{eq:eqclass}
\vec x \in [\widetilde \x]_f^{\circ}   \Longleftrightarrow 
W_{\widetilde \x}^{(k)} \vec x + \vec b_{\widetilde \x}^{(k)} > 0, \,\, k = 1, \ldots, L,
\end{eqnarray}
i.e., $[\widetilde \x]_f^{\circ} = P_{\widetilde \x}^{\circ}$ and $\overline{[\widetilde \x]}_f= P_{\widetilde \x}$.

\subsection{Third Part of Proof of Theorem~\ref{thm:duality}}
From the fact that the equivalence classes are finite intersections of either closed or open half-spaces it follows that an equivalence class $[\textbf{x}]_f$ is convex. Further, observe that the bundle of hyperplanes that generate the corresponding equivalence classes establishes a tessellation of the input space.
Now, consider the relative interior $relint(F)$ of a subface $F$ of $\overline{[\x]}_f$ that has non-empty intersection with $[\x]_f$. 
Suppose that there are two points $\z_1, \z_2 \in relint(F)$ with $\z_1 \sim \textbf{x}$ and $\z_2 \not \sim \x$.
This means that there is a hyperplane (in the bundle of hyperplanes of the tessellation) that separates $\z_1$ from $\z_2$, thus splitting $[\x]_f$ into two non-empty parts corresponding to different codes, which contradicts the construction principle of the equivalence class $[\x]_f$. Consequently, we have that
\begin{equation}
\label{eq:relintProp}
relint(F)  \cap [\x]_f \neq \emptyset \Longrightarrow relint(F) \subseteq [\x]_f.
\end{equation}
Now, consider for a point $\z \in [\x]_f$ the least upper subface $F$ of $\overline{[\textbf{x}]}_f$ with 
$\z \in F$. We prove that $\z \in relint(F)$ which is trivial of a corner point. Therefore, suppose that $F$ is not a corner point (dimension 0), and let us suppose the contrary that $\z \not \in relint(F)$. 
This means $\z \in \partial(F)=F\backslash relint(F)$. From $dim(\partial(F))< dim(F)$ it follows that
$F$ is not a least upper subface, which is a contradiction to the assumption.

Therefore, we obtain the result:
\begin{lemma}
\label{lemma:relint}
Let $\z \in [\x]_f$ and let $F_z$ be the least upper subface with $\z \in F_{\z} \subseteq [\x]_f$.
Then, $\z \in relint(F_{\z})$.
\end{lemma}
As a consequence $[\x]_f$ can be constructed as disjoint union of relative interiors of subfaces of the induced 
polyhedron $\overline{[\x]}_f$. The lattice structure of 
$$
\left(\{relint(F) \subseteq [\x]_f\mid F \,\mbox{subface of }[\x]_f \}, \prec\right)
$$ 
is inherited from the lattice structure of subfaces of a polyhedron
$$
(\{ F \subseteq [\x]_f\mid F \,\mbox{subface of }[\x]_f \}, \subseteq )
$$ 
by defining the partial ordering $relint(F_1) \prec relint(F_2)$ if and only if
$F_1 \subseteq  F_2$.
\hfill\qed

\section{Appendix}
\label{appendix_B}

\subsection{Experimental set up}
In this section we present full experimental setup supporting our theoretical foundings. We shortly introduce the reader to exact software and hardware specifics, as well as we detail the simulation's results.

All experiments were performed on NVIDIA DGX-1 station (CPU: Intel Xeon E5-2698 v4 2,2 GHz, 20-Core; GPU: 4x Tesla V100, 64GB; OS: DGX Base OS, 4.0.5, Ubuntu 18.04.2) in single GPU mode. 
Implementation is based on TensorFlow v.1.13.1, random seeds for Python environment, NumPy library and TensorFlow library were set to value of 1234. All layers in the described below architectures were initialized with Xavier kernel (seed=0) and zeros bias initializers. 

For our experiments we used two architectures: Autoencoder and VGG16. Since our interest lies in investigation of fully connected(FC) layers, we kept original VGG16 backbone for all our experiments and changed only number of layers and amount of nodes. We picked 512 nodes as a basis for our experiments since it is commonly used solution for MNIST and CIFAR10 datasets in the community. With that we had following architectures in our set up: three layer Autoencoder with 128, 64, 32 nodes respectively, one layer VGG16 with 512 nodes, two layer VGG16 with 256 nodes in each layer, and three layer VGG16 with 256, 128, 128 nodes respectively. The overview of architectures used in experiments is provided in Table~\ref{tab:arch_sum}. Although, it is important to mention, that we did not use dropout in our experiments, since this regularization technique is not covered in our theory.
As for the learning procedure for all VGG16 experiment we set batch size to 256, with softmax loss and gradient descent optimizer. 

\begin{table}[h]
\centering
\begin{tabular}{|c|c|c|c|c|c|c|c|}
\hline
\multicolumn{1}{|l|}{} & \textbf{Autoencoder} & \multicolumn{3}{c|}{\textbf{VGG16}} \\ \hline
\textbf{FC-0} & 128 & 512 & 256 & 256 \\ \hline
\textbf{FC-1} & 64  & - & 256 & 128 \\ \hline
\textbf{FC-2} & 32  & - & - & 128 \\ \hline
output & decoder & softmax & softmax & softmax \\ \hline
\end{tabular}
\caption{Summary of experimental architectures}
\label{tab:arch_sum}
\end{table}

As for the learning procedure for all VGG16 experiment we set batch size to 256, with softmax loss and gradient descent optimizer. 

\subsection{Experiments on Autoencoder}
\label{ss:autoenc_exp}

\begin{figure}[H]
    \centering 
  \includegraphics[width=0.85\linewidth, scale=0.7]{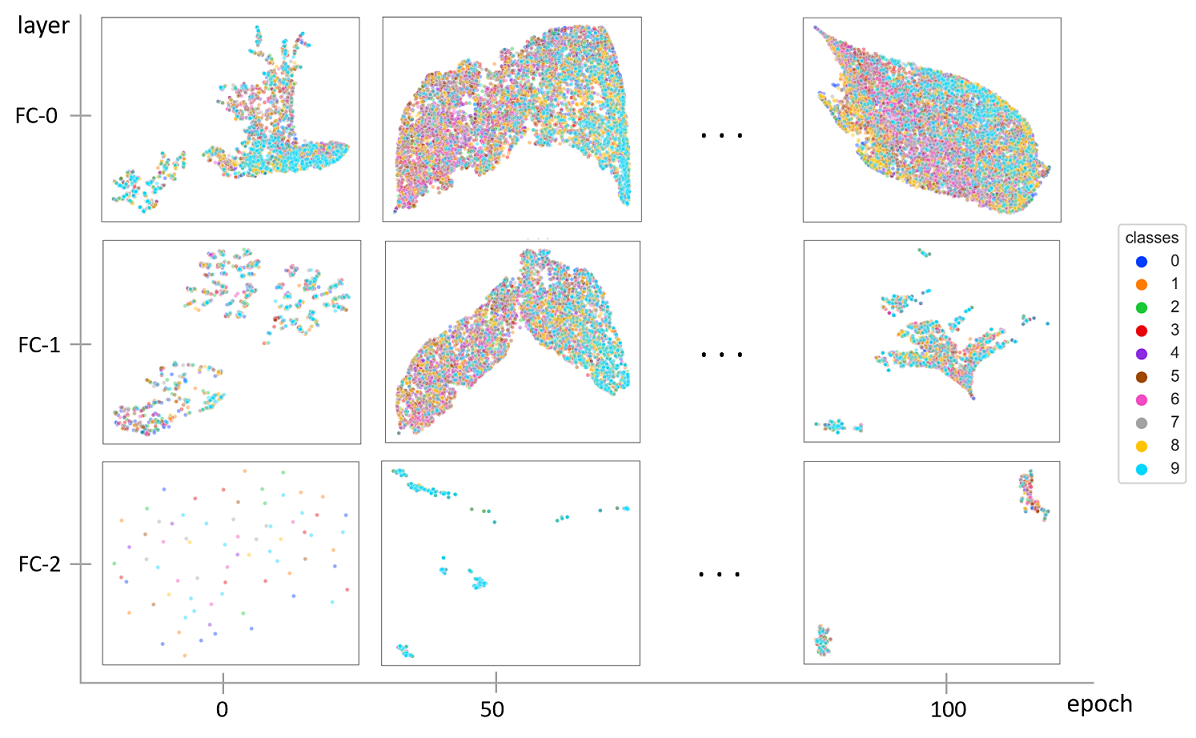}
\caption{Autoencoder network trained on CIFAR10. Rows show changes of the clusters of ReLU codes over the training time, columns show corresponding changes withing the network layers.}
\label{fig:autornc_cifar}
\end{figure}

We trained Autoencoder for two datasets MNIST shown in Figure~\ref{fig:ae} and CIFAR10 shown in Figure~\ref{fig:autornc_cifar}. As we can see for both datasets features become more invariant with depth of the Autoencoder, as we did assume.

In addition to this experiment we trained same architecture only on two classes with the lowest separability from each dataset. For MNIST that are labels 4 and 9, which are noted on Figure~\ref{fig:ae_mnist_2cl} as labels 0 and 1. For CIFAR10 such classes are frog and dear which are represented on Figure~\ref{fig:ae_cifar_2cl} as 0 and 1.

\begin{figure}[H]
    \centering 
\begin{subfigure}{0.48\textwidth}
  \includegraphics[width=\linewidth]{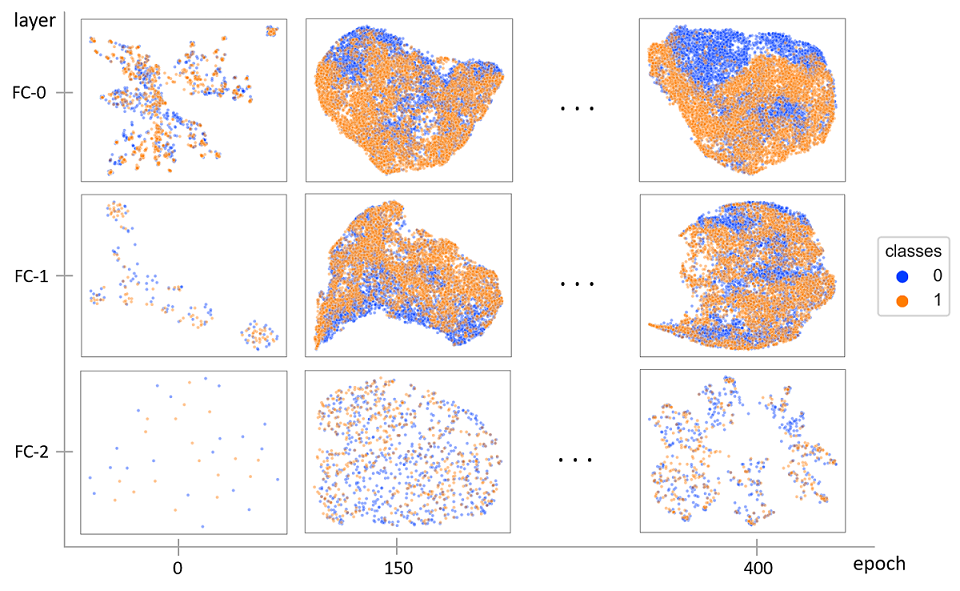}
  \caption{MNIST}
  \label{fig:ae_mnist_2cl}
\end{subfigure}\hfil 
\begin{subfigure}{0.48\textwidth}
  \includegraphics[width=\linewidth]{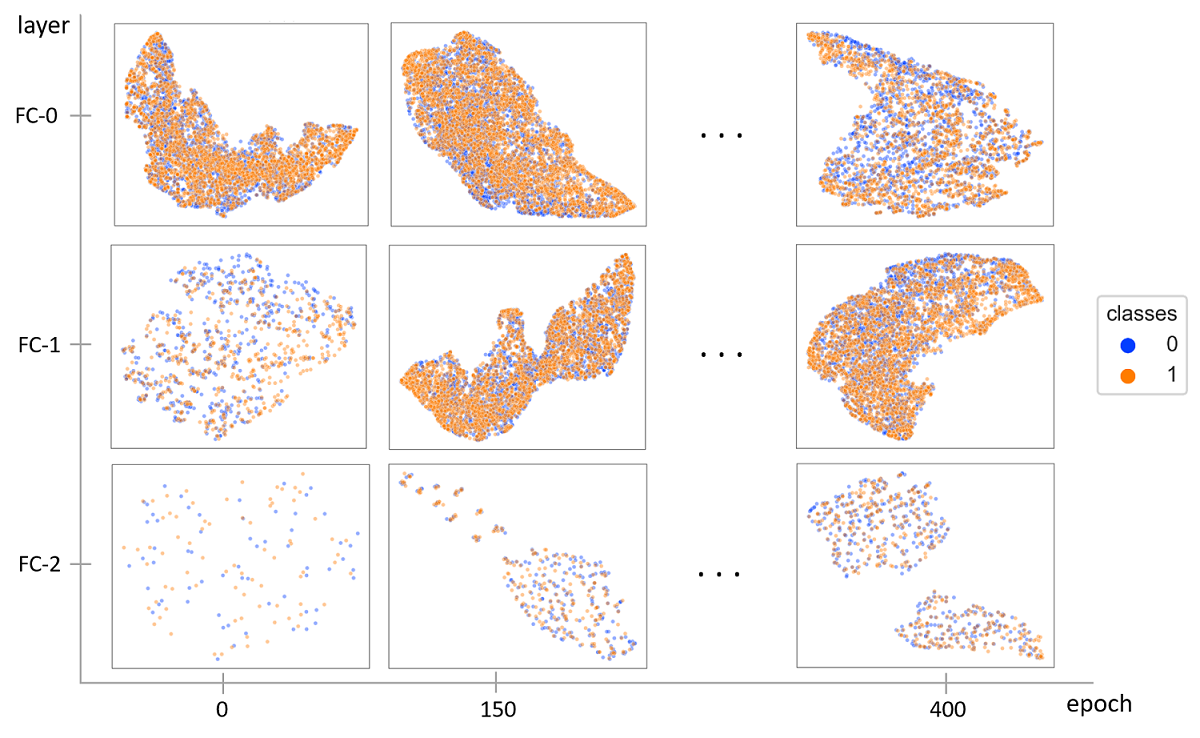}
  \caption{CIFAR10}
  \label{fig:ae_cifar_2cl}
\end{subfigure}\hfil 
\caption{Autoencoder network trained on two selected classes from MNIST and CIFAR10 datasets. Rows show changes of the clusters of ReLU codes over the training time, columns show corresponding changes withing the network layers.}
\label{fig:cluster}
\end{figure}

\subsection{Experiments on VGG16}
\label{ss:vgg16_exp}

\begin{figure}[H]
    \centering 
\begin{subfigure}{0.45\textwidth}
  \includegraphics[width=\linewidth]{results/exp/vgg16_mnist.png}
  \label{fig:ae_mnist_2cl}
\end{subfigure}\hfil 
\begin{subfigure}{0.45\textwidth}
  \includegraphics[width=\linewidth]{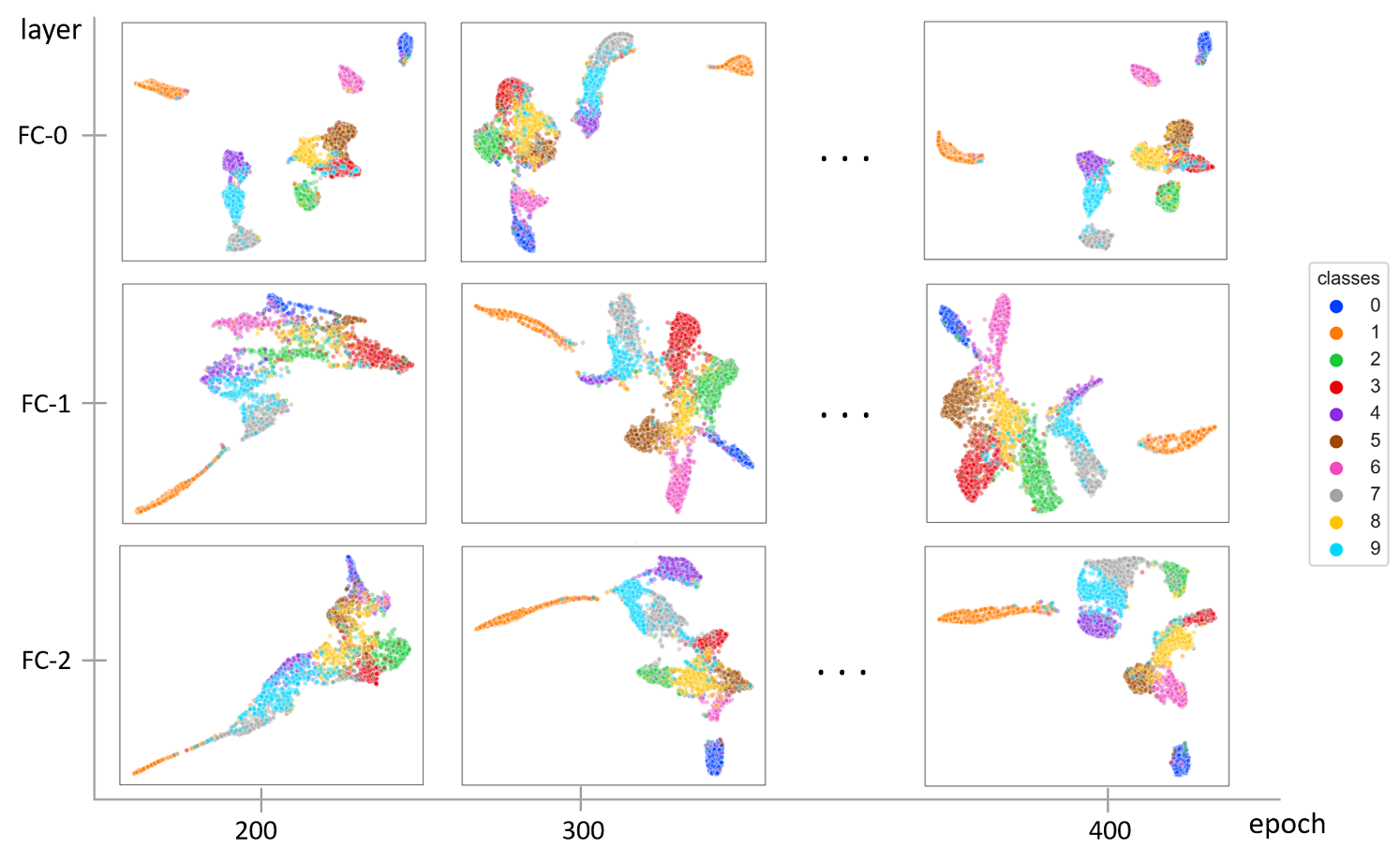}
  \label{fig:ae_cifar_2cl}
\end{subfigure}\hfil 

\medskip

\begin{subfigure}{0.45\textwidth}
  \includegraphics[width=\linewidth]{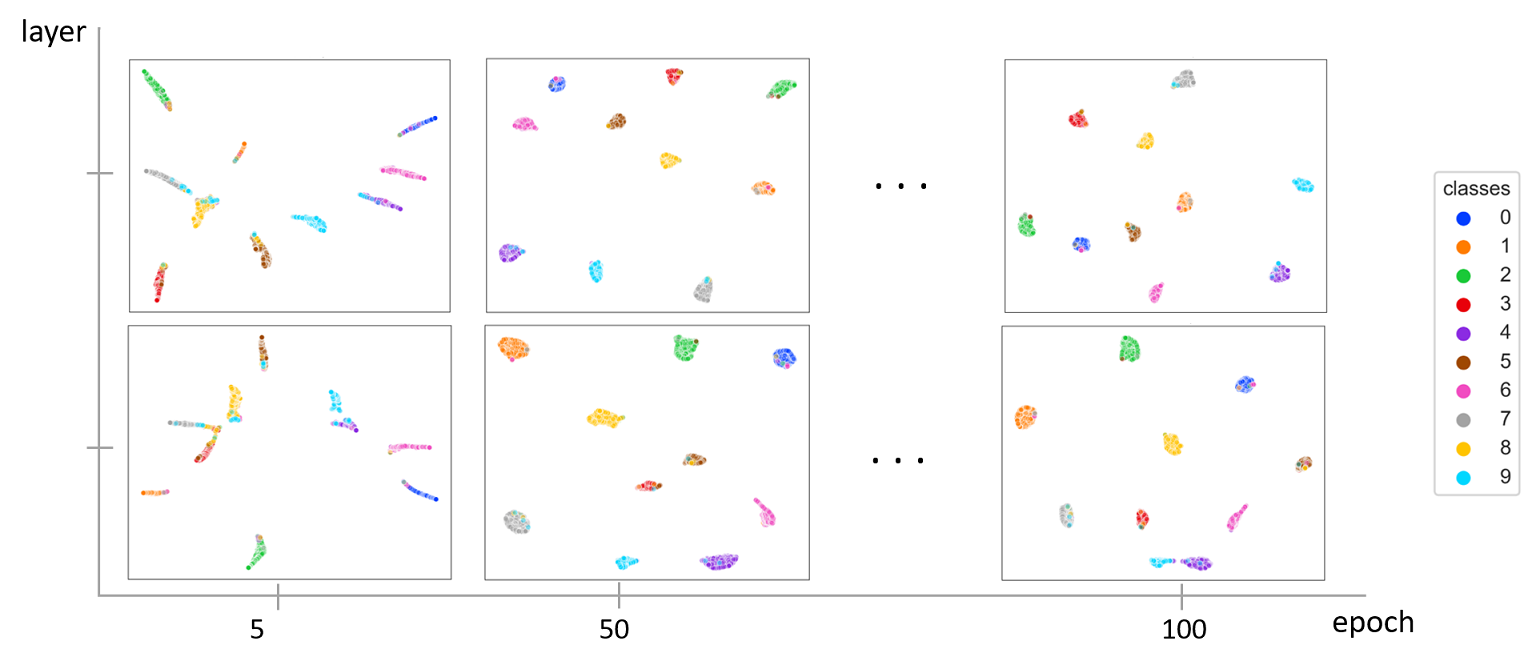}
  \label{fig:ae_mnist_2cl}
\end{subfigure}\hfil 
\begin{subfigure}{0.45\textwidth}
  \includegraphics[width=\linewidth]{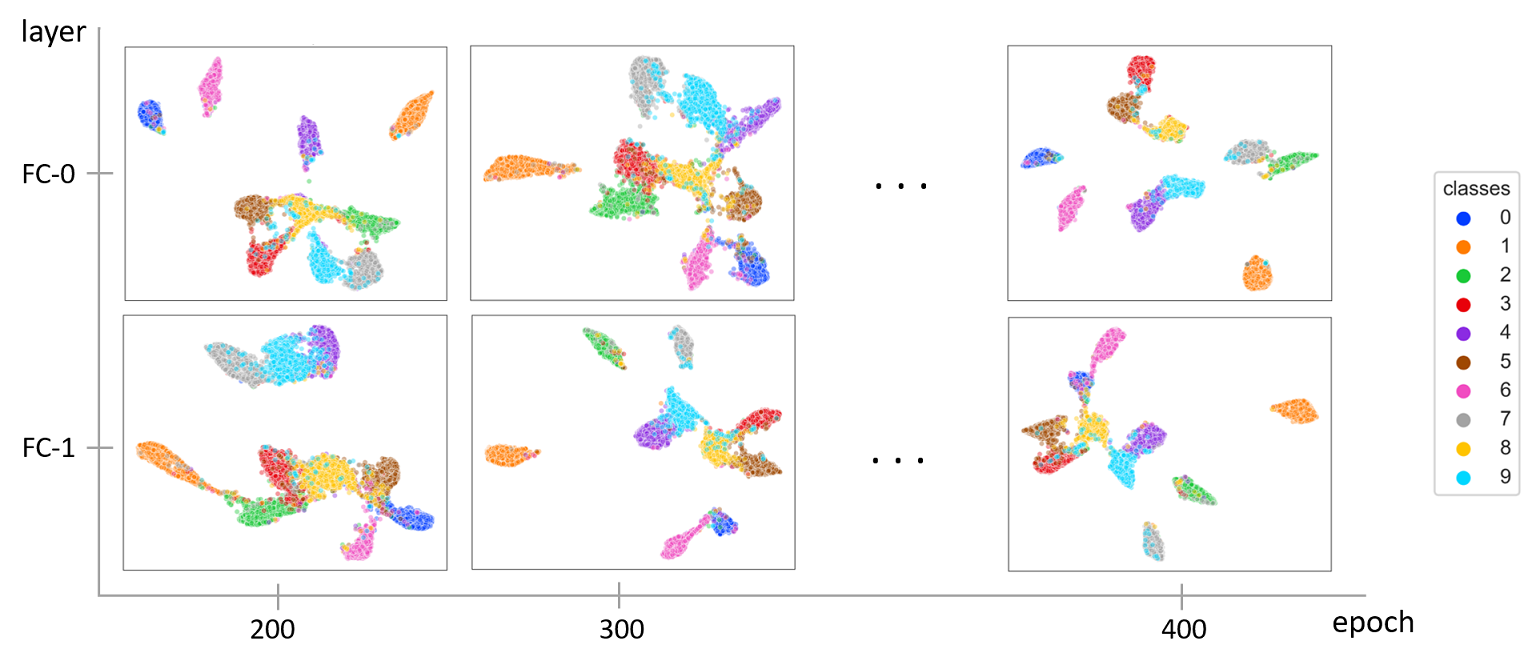}
  \label{fig:ae_cifar_2cl}
\end{subfigure}\hfil 

\medskip

\begin{subfigure}{0.45\textwidth}
  \includegraphics[width=\linewidth]{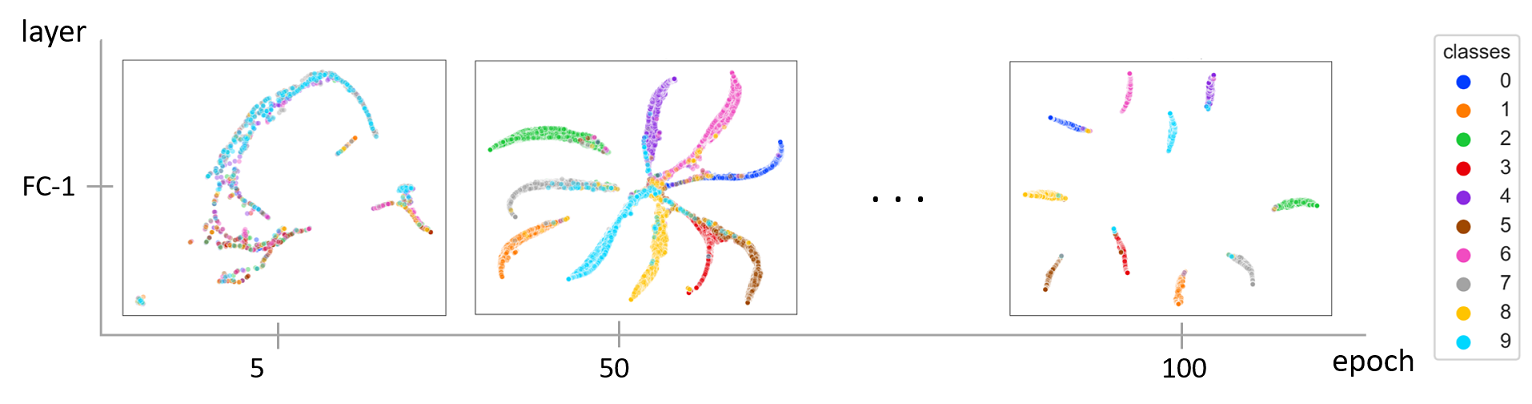}
  \caption{learning rate 0.1}
  \label{fig:ae_mnist_2cl}
\end{subfigure}\hfil 
\begin{subfigure}{0.45\textwidth}
  \includegraphics[width=\linewidth]{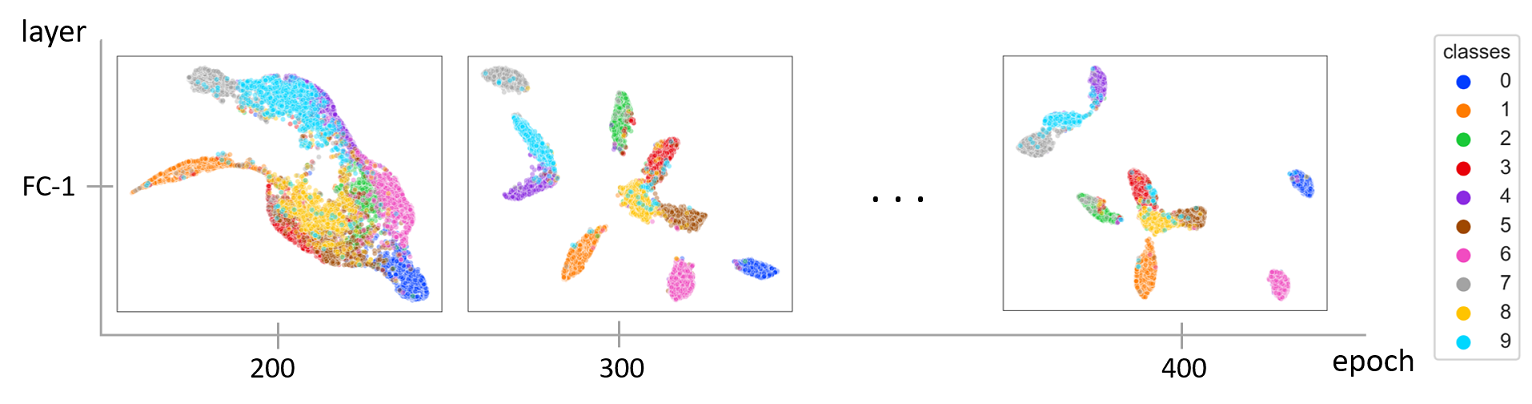}
  \caption{learning rate 0.001}
  \label{fig:ae_cifar_2cl}
\end{subfigure}\hfil 

\caption{VGG16 network trained on MNIST with different learning rates, a) 0.1 and b) 0.001.}
\label{fig:vgg16_mnist_full}
\end{figure}

\begin{figure}[H]
    \centering 
\begin{subfigure}{0.45\textwidth}
  \includegraphics[width=\linewidth]{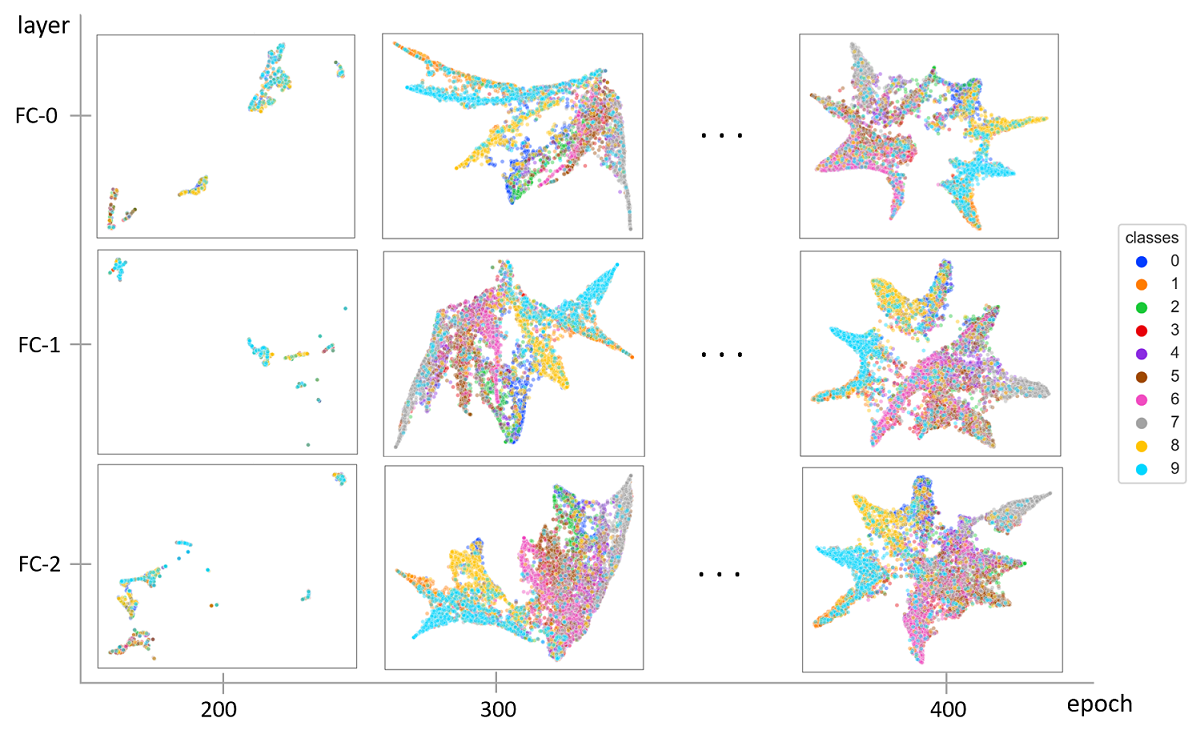}
  \label{fig:ae_mnist_2cl}
\end{subfigure}\hfil 
\begin{subfigure}{0.45\textwidth}
  \includegraphics[width=\linewidth]{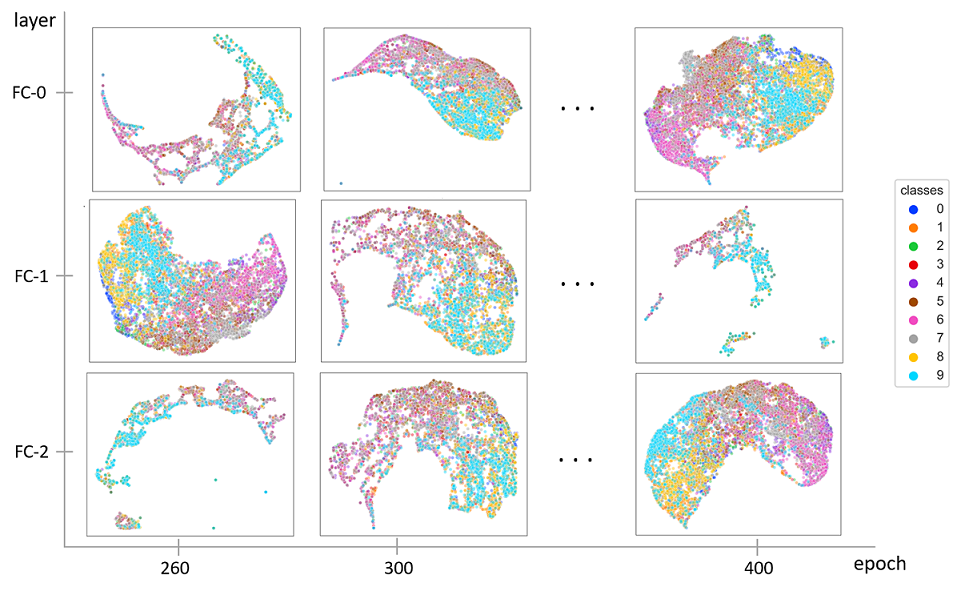}
  \label{fig:ae_cifar_2cl}
\end{subfigure}\hfil 

\medskip

\begin{subfigure}{0.45\textwidth}
  \includegraphics[width=\linewidth]{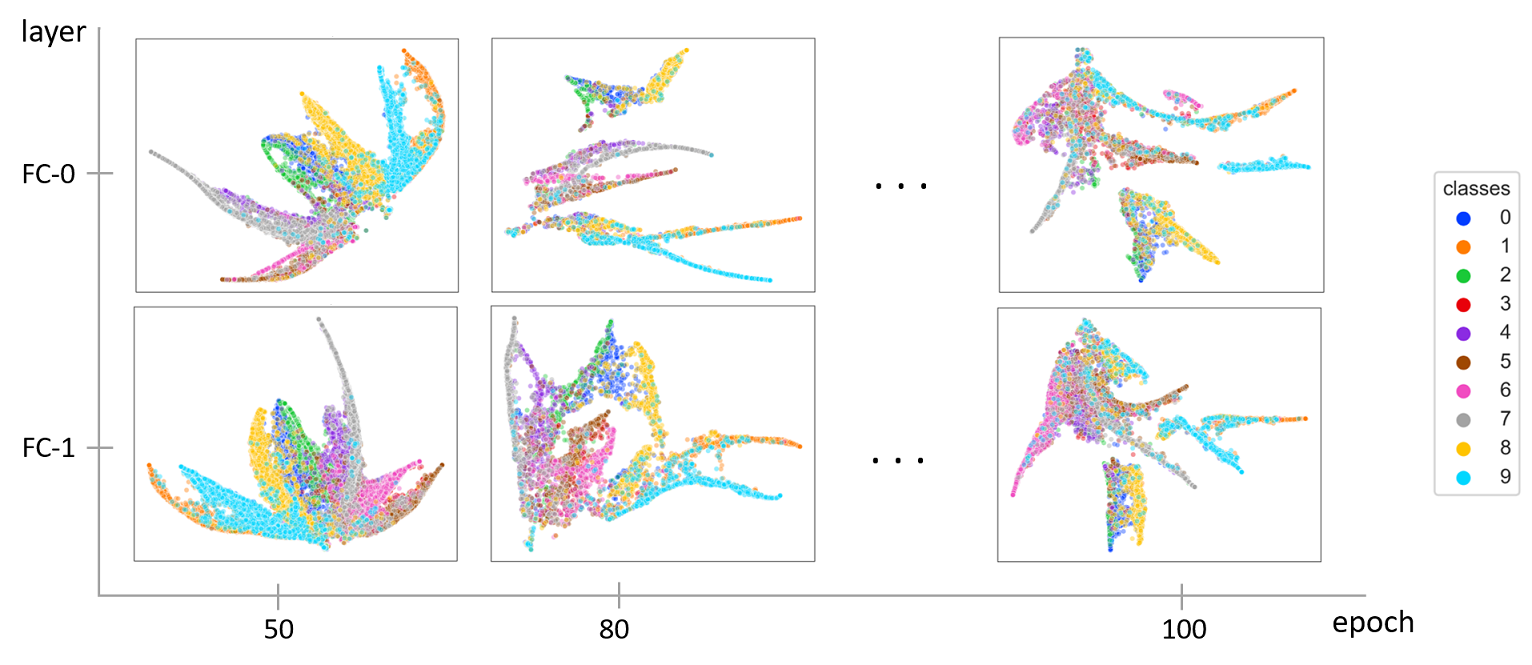}
  \label{fig:ae_mnist_2cl}
\end{subfigure}\hfil 
\begin{subfigure}{0.45\textwidth}
  \includegraphics[width=\linewidth]{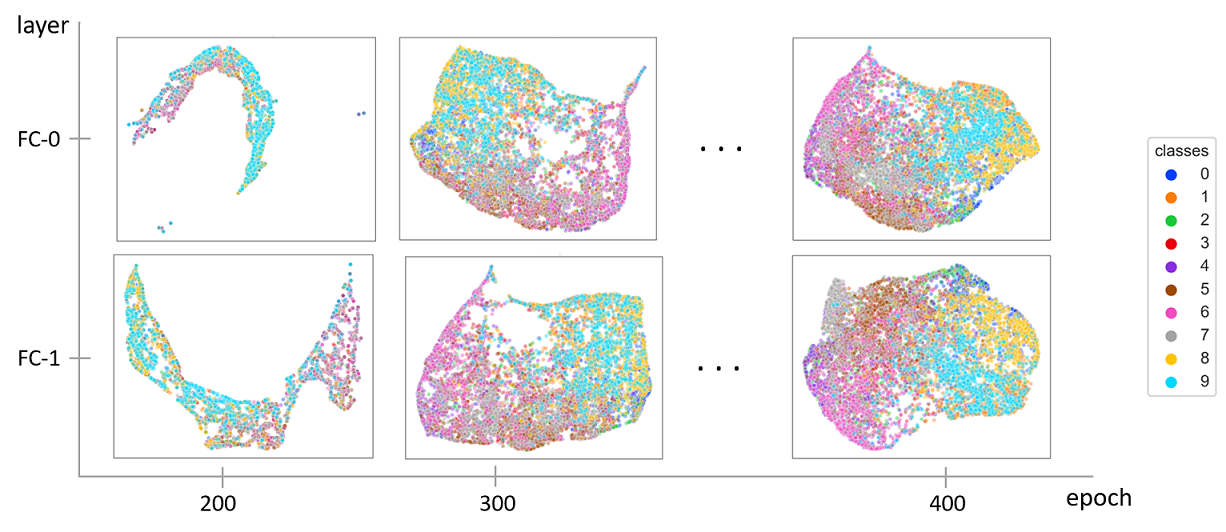}
  \label{fig:ae_cifar_2cl}
\end{subfigure}\hfil 

\medskip

\begin{subfigure}{0.45\textwidth}
  \includegraphics[width=\linewidth]{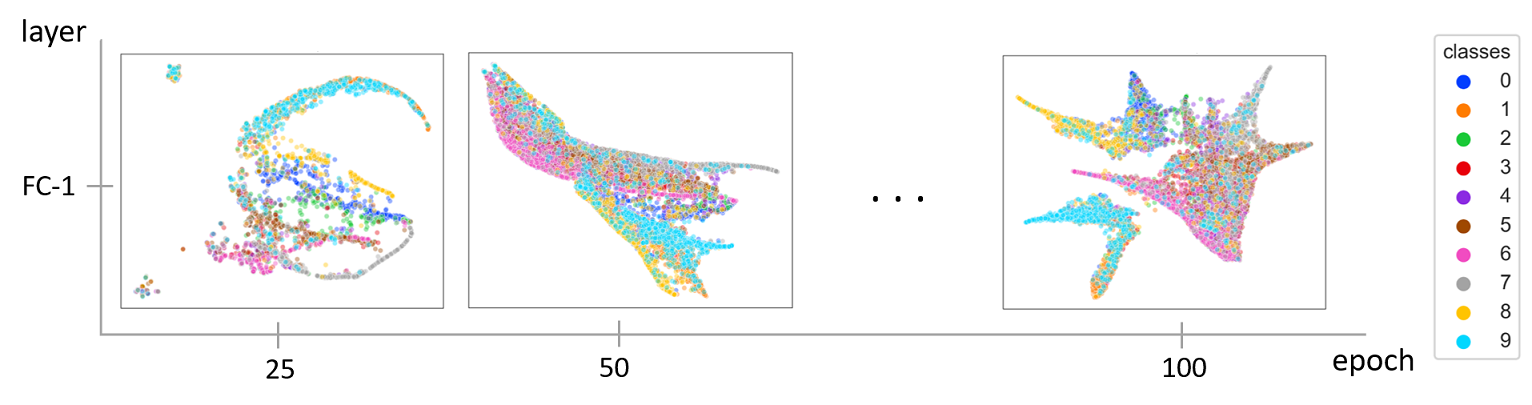}
  \caption{learning rate 0.1}
  \label{fig:ae_mnist_2cl}
\end{subfigure}\hfil 
\begin{subfigure}{0.45\textwidth}
  \includegraphics[width=\linewidth]{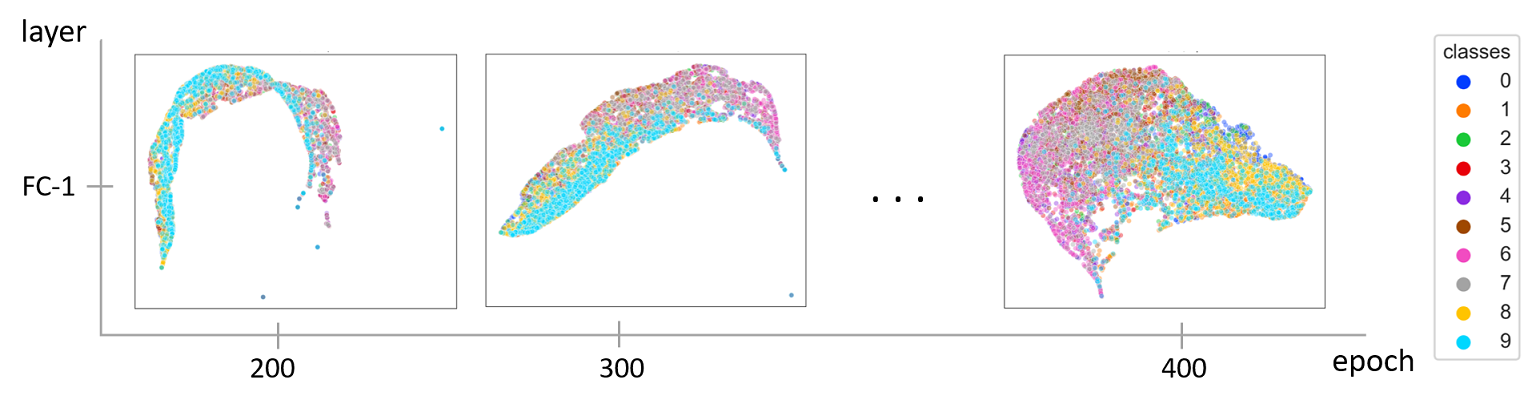}
  \caption{learning rate 0.001}
  \label{fig:ae_cifar_2cl}
\end{subfigure}\hfil 

\caption{VGG16 network trained on CIFAR10 with different learning rates, a) 0.1 and b) 0.001.}
\label{fig:vgg16_cifar_full}
\end{figure}

In further experiments we trained VGG16 on MNIST and CIFAR10 datasets. Each VGG16 experiment described in Table~\ref{tab:arch_sum} for each dataset we repeated twice: for learning rate 0.1 and 0.001. The resulting clustering of ReLU codes is shown in Figure~\ref{fig:vgg16_mnist_full} and Figure~\ref{fig:vgg16_cifar_full}.

\end{document}